\documentclass{article}

\usepackage{arxiv}

\usepackage[utf8]{inputenc} 
\usepackage[T1]{fontenc}    
\usepackage{url}            
\usepackage{booktabs}       
\usepackage{amsfonts}       
\usepackage{nicefrac}       
\usepackage{microtype}      
\usepackage{lipsum}
\usepackage{graphicx}
\graphicspath{ {./images/} }
\usepackage{array}
\usepackage{amsmath}
\usepackage{float}
\usepackage{listings }
\usepackage{subcaption}
\usepackage{multirow}

\usepackage[frozencache,cachedir=.]{minted}
\usepackage{hyperref}       

\title{tsBNgen: A Python Library to Generate Time Series Data from an Arbitrary Dynamic Bayesian Network Structure }

\author{
  Manie Tadayon\\
  Department of Electrical and Computer Engineering\\
  UCLA University\\
  Los Angeles, CA  \\
  \texttt{manitadayon@ucla.edu} \\
   \And
  Greg Pottie \\
  Department of Electrical and Computer Engineering\\
  UCLA University\\
  Los Angeles, CA \\
  \texttt{pottie@ee.ucla.edu} \\
}
\begin{document}
\maketitle
\begin{abstract}
Synthetic data is widely used in various domains. This is because many modern algorithms require lots of data for efficient training, and data collection and labeling usually are a time-consuming process and are prone to errors. Furthermore, some real-world data, due to its nature, is confidential and cannot be shared. 
Bayesian networks are a type of probabilistic graphical model widely used to model the uncertainties in real-world processes. Dynamic Bayesian networks are a special class of Bayesian networks that model temporal and time series data.
In this paper, we introduce the tsBNgen, a Python library to generate time series and sequential data based on an arbitrary dynamic Bayesian network. The package, documentation, and examples can be downloaded from https://github.com/manitadayon/tsBNgen (\href{https://github.com/manitadayon/tsBNgen}{tsBNgen}).

\end{abstract}


\section{Introduction}
\label{sec:headings}
Real-world data collection and labeling are time-consuming and error-prone tasks that can take up to years to finish. Furthermore, some real-world data, due to its confidential nature, cannot be distributed or analyzed. Moreover, many modern algorithms, such as neural networks, require a large amount of data for efficient training. For example, in \cite{tadayon2020comparative}, the authors show that under limited data, a more classical algorithm such as a hidden Markov model (HMM) can outperform the long short term memory (LSTM) algorithm. 

Bayesian networks (BN), have been used extensively in modeling various applications, such as \cite{vomlel2004bayesian} and \cite{mouri2016bayesian}. A dynamic Bayesian network (DBN) is a variant of a BN used to model temporal and time series data. Well-known examples of DBNs in practice are HMMs, conditional random field (CRFs), and their variants.
For example; in \cite{tadayon2020predicting}, the authors used an HMM to predict student performance in an educational video game. They used a discrete HMM to measure student mastery of concepts as they go through levels of the game. In \cite{xia2020sentiment}, the authors proposed an approach for online review sentiment classification and used the CRF algorithm to extract the emotional characteristics from fragments of the review. 

Scientists and researchers have proposed various methods to generate synthetic data such as \cite{frid2018synthetic}, \cite{bowles2018gan} and \cite{esteban2017real} using generative adversarial network (GAN). In \cite{esteban2017real}, the authors proposed a Recurrent GAN (RGAN) and Recurrent Conditional GAN (RCGAN) to produce realistic real-valued multi-dimensional time series data. Although the GAN has numerous image processing and computer vision applications and produces satisfactory results, it suffers from the following limitations: 1- It is hard to train 2- it is sometimes unstable and might not never converge. 3- It is significantly harder to train for text than images. 4- It generally requires lots of data for training and might not be a good choice when there is limited or no available data. 

In this paper, we introduce the \href{https://github.com/manitadayon/tsBNgen}{tsBNgen}, a Python library to generate time series and sequential data from an arbitrary dynamic Bayesian network structure. tsBNgen is available to download at https://github.com/manitadayon/tsBNgen. The rest of this paper is organized as follow. Section 2 reviews the features and capabilities of this software. Section 3 reviews the instructions on how to use the software, followed by an example. Section 4 concludes the paper.

\section{Features and Capabilities }
\label{sec:Algorithm}

tsBNgen is an open-source Python library released under MIT license. It generates time series data corresponding to any Bayesian network structures. The following is the list of supported features and capabilities of the tsBNgen.
\begin{itemize}
    \item Easy and simple interface.
    \item Support for discrete, continuous, and hybrid networks (a mixture of discrete and continuous nodes).
    \item Supports an arbitrary number of nodes with any interconnections.
    \item Use multinomial distributions for discrete and Gaussian distributions for continuous nodes.
    \item Supports arbitrary loopback values for temporal dependencies.
    \item Easy to modify and extend the code to support for the new structure. 
\end{itemize}

Loopback is defined as the length of temporal dependency. For example, loopback of one means node at time $t$ is connected to some nodes at time $t$+1.

\section{Instruction}
To use tsBNgen, either clone this repository \href{https://github.com/manitadayon/tsBNgen}{tsBNgen} or install the software using the following commands:

\begin{minted}{python}
pip install tsBNgen
\end{minted}

After the software is successfully installed. Import necessary libraries and functions as follows.
\begin{minted}{python}
from tsBNgen import *
from tsBNgen.tsBNgen import * 
\end{minted}

The above two lines import all the functions for tsBNgen. Next, we review an example of how to use tsBNgen to generate an arbitrary DBN structure. Before going over an example, we need to define some notations.

$T$ = Length of each time series.

$D$ = Dimension of time series.

$N$ = Number of samples (time series)

$u_{i0},u_{i1},...u_{iT}$ = sequence of node i.

$N{ui}$ = Number of possible symbols for node i (if it is discrete).

\subsection{\textbf{Example }}

This Example refers to the figure 1, which is the diagram of HMM in which $u_{00}, u_{01},..., u_{0T}$ refer to the state sequence and $u_{10}, u_{11},..., u_{0T}$ refer to the observations. The following table summarizes the setting and configuration corresponding to this model.

\begin{table}[htb]
\centering
\caption{Parameter Setting for Example 1}
\resizebox{0.45\textwidth}{!}{
\begin{tabular}{ |c|c| } 
\hline
Architecture & Value \\
\hline
$T$ & 20 \\
\hline
$N$ & 1000\\
\hline
Loopback & 1 \\
\hline
\# of discrete nodes per time step & 1 \\
\hline
\# of possible levels for node $u_{0}$ per time point & 4 \\
\hline
\# of continuous nodes per time step & 1 \\
\hline
\end{tabular}
}
\end{table}

\begin{figure}[htb]
    \centering
    \includegraphics[scale=0.85]{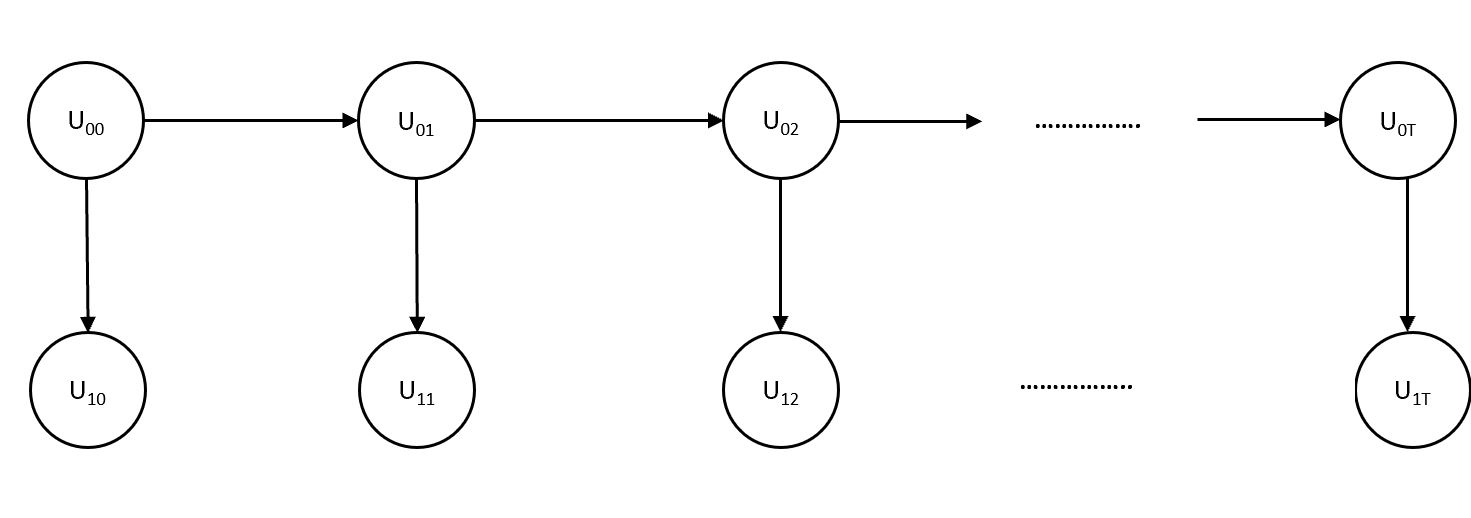}
    \caption{Synthetic Time Series Under Example 1}
    \label{fig:label1}
\end{figure}

The conditional probability distributions (CPDs) for this network are listed in Tables 2, 3, and 4.

\begin{table}[t]
        \scalebox{1.0}{
        \begin{minipage}[h]{0.3\textwidth}
        \caption{\textbf{CPD for Initial Time}}
            \centering
            \resizebox{0.40\columnwidth}{!}{
            \begin{tabular}{|c|c|}
            \hline
            $u_{00}$ & $P(u_{00})$ \\
            \hline
                1 & 0.25  \\
                \hline
                2 & 0.25  \\
                 \hline
                3 & 0.25  \\
                 \hline
                4 & 0.25  \\
                 \hline
            \end{tabular}
            }
        \end{minipage}
        \begin{minipage}[h]{0.4\textwidth}
        \caption{\textbf{CPD for Time Step $t_{1}$ to $t_{T}$}}
            \centering
            \begin{tabular}{|c|c|c|}
             \hline
            $u_{0t-1}$ & $u_{0t}$ &  $P(u_{0t}| u_{0t-1})$ \\
            \hline
                1 & 1 & 0.6 \\
                \hline
                2 & 1 & 0.3\\
                \hline
                3 & 1 & 0.05\\
                \hline
                4 & 1 & 0.05\\
                \hline
                1 & 2 & 0.25\\
                \hline
                2 & 2 & 0.4\\
                \hline
                3 & 2 & 0.25\\
                \hline
                4 & 2 & 0.1\\
                \hline
                1 & 3 & 0.1 \\
                \hline
                2 & 3 & 0.3\\
                \hline
                3 & 3 & 0.4\\
                \hline
                4 & 3 & 0.2\\
                \hline
                1 & 4 & 0.05\\
                \hline
                2 & 4 & 0.05\\
                \hline
                3 & 4 & 0.4\\
                \hline
                4 & 4 & 0.5\\
                \hline
            \end{tabular}
            
        \end{minipage}
        \begin{minipage}[h]{0.4\textwidth}
        \caption{\textbf{CPD for the Continuous Node}}
            \centering
            \resizebox{0.40\columnwidth}{!}{
            \begin{tabular}{|c|c|c|}
            \hline
            $u_{1t}$ & $\mu$ & $\sigma$ \\
            \hline
                1 & 20 & 5  \\
                \hline
                2 & 40 & 5  \\
                 \hline
                3 & 60 & 5  \\
                 \hline
                4 & 80 & 5  \\
                 \hline
            \end{tabular}
            }
        \end{minipage}
        }
    \end{table}
The code to model and generate data for figure 1 is as follows:
\begin{minted}{python}
import time
START=time.time()
T=20
N=1000
N_level=[4]
Mat=pd.DataFrame(np.array(([0,1],[0,0]))) # HMM
Node_Type=['D','C']

CPD={'0':[0.25,0.25,0.25,0.25],'01':{'mu0':20,'sigma0':5,'mu1':40,'sigma1':5,
    'mu2':60,'sigma2':5,'mu3':80,'sigma3':5}}

Parent={'0':[],'1':[0]}


CPD2={'00':[[0.6,0.3,0.05,0.05],[0.25,0.4,0.25,0.1],[0.1,0.3,0.4,0.2],
[0.05,0.05,0.4,0.5]],'01':{'mu0':20,'sigma0':5,'mu1':40,'sigma1':5,
    'mu2':60,'sigma2':5,'mu3':80,'sigma3':5
}}

loopbacks={'00':[1]}

Parent2={'0':[0],'1':[0]}


Time_series1=tsBNgen(T,N,N_level,Mat,Node_Type,CPD,Parent,CPD2,Parent2,loopbacks)

Time_series1.BN_data_gen() 
FINISH=time.time()
print('Total Time is',FINISH-START)
\end{minted}

The total time is 2.06 (s), and running the model through the HMM algorithm gives us more than 93.00 \% accuracy for even five samples.

\section{Conclusion}

In this paper, we introduced the tsBNgen, a python library to generate synthetic data from an arbitrary BN. We discussed the features and capabilities of our software. We discussed how to use the software using an example.  
For more examples, up-to-date documentation, and instructions, please visit the GitHub page at https://github.com/manitadayon/tsBNgen.

\bibliographystyle{unsrt}  
\bibliography{references}  






\end{document}